\documentclass[11pt]{article}
\usepackage{amsmath}
\usepackage{amssymb}
\usepackage[ruled,vlined,linesnumbered]{algorithm2e}
\DontPrintSemicolon
\SetKwInput{KwRequire}{Require}
\SetKwInput{KwEnsure}{Ensure}
\usepackage{parskip} 
\usepackage{arydshln}
\usepackage[table]{xcolor}
\usepackage{booktabs}
\usepackage{multirow}
\usepackage[final]{acl}

\usepackage{times}
\usepackage{latexsym}

\usepackage[T1]{fontenc}

\usepackage[utf8]{inputenc}

\usepackage{microtype}

\usepackage{inconsolata}

\usepackage{graphicx}

\title{Uncovering and Mitigating Aggregation-Induced Reward Hacking in Multi-Reward Reinforcement Learning}

\author{
\textbf{Yu Yuan}\textsuperscript{1,2},
\textbf{Yaoyou Fan}\textsuperscript{3},
\textbf{Lili Zhao}\textsuperscript{4,$\dagger$},
\textbf{Guangting Zheng}\textsuperscript{1}, \\
\textbf{Kai Zhang}\textsuperscript{1,2},
\textbf{Lu Pan}\textsuperscript{4},
\textbf{Ke Zeng}\textsuperscript{4},
\textbf{Qi Liu}\textsuperscript{1, 2, $\dagger$}\\
\textsuperscript{1}University of Science and Technology of China \\
\textsuperscript{2}State Key Laboratory of Cognitive Intelligence \\
\textsuperscript{3}Peking University \hspace{0.1em}
\textsuperscript{4}Longcat-Interaction Team, Meituan \\
\texttt{yyhappier@mail.ustc.edu.cn},
\hspace{0.1em}
\texttt{zhaolili08@meituan.com},
\hspace{0.1em}
\texttt{qiliuql@ustc.edu.cn}
}

\begin{document}
\maketitle

\begingroup
\renewcommand{\thefootnote}{\fnsymbol{footnote}}
\footnotetext[2]{\ Corresponding authors.}
\endgroup

\begin{abstract}

Reinforcement learning fine-tuning of large language models increasingly adopts multiple reward dimensions, including verifiable rules, task-specific evaluators, and learned reward models, to provide richer supervision across diverse capabilities.
These dimensions are commonly scalarized with fixed aggregation weights. 
We identify a failure mode in which aggregation itself induces reward hacking: static projection aliases qualitatively different reward profiles into a single scalar, steering optimization toward whichever dimensions are easiest, densest, or systematically favored by the reward signal. Over training, this traps the policy in suboptimal profiles and prevents convergence to better-balanced ones that would yield higher task performance. 
To address this, we propose \textbf{A}daptive \textbf{M}ulti-\textbf{R}eward \textbf{P}rojection (\textbf{AMRP}), a lightweight online method that reallocates aggregation weights using three signals—relative shortfall, reward volatility, and recent progress—increasing pressure on lagging, unstable, or stagnant dimensions while relieving saturated ones.  
Across structured reasoning, citation-grounded generation, and open-ended alignment under GRPO, AMRP consistently improves reward-profile balance and downstream performance over fixed and dynamic weighting baselines; it also remains effective with GDPO and PPO, supporting compatibility across RL algorithms.
Our code is available at \url{https://github.com/yyhappier/AMRP.git}.

\end{abstract}

\begin{figure}[t]
\centering
\includegraphics[width=0.95\linewidth]{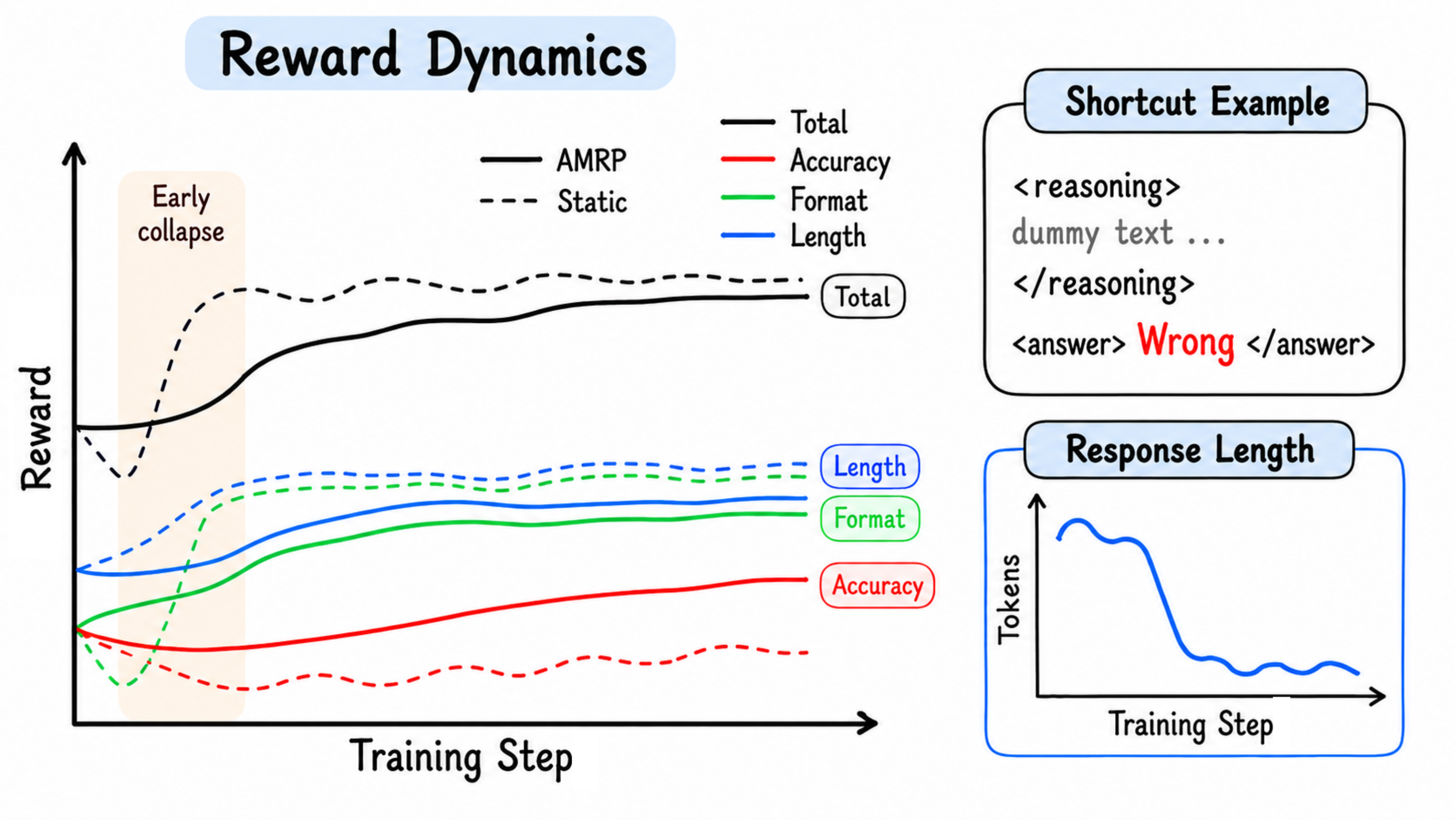}
\caption{
Overview of aggregation-induced reward hacking and AMRP in mathematical reasoning.
The reward profile \(\mathbf r_{\mathrm{math}}=[r_{\mathrm{acc}},r_{\mathrm{fmt}},r_{\mathrm{len}}]\) measures answer accuracy, format compliance, and length control.
Static aggregation can increase scalar reward by rapidly improving format and length while accuracy remains low, leading to shortcut rollouts.
AMRP mitigates this failure by adapting projection weights online.
}
\label{fig:intro-collapse}
\end{figure}

\section{Introduction}

Post-training has become standard for adapting Large Language Models to follow instructions and solve complex tasks~\citep{ouyang2022training,guo2025deepseek}, with  
many pipelines now optimizing multiple reward dimensions simultaneously—for example, answer correctness, format compliance, and code execution. A common practice is to aggregate these into a single scalar with fixed weights throughout training~\citep{hayes2022practical}, a design that is simple and compatible with standard RL optimizers such as PPO~\citep{schulman2017proximal} and GRPO~\citep{shao2024deepseekmath}. However, this projection is lossy: qualitatively different reward profiles can yield the same scalar, obscuring which dimensions actually drive learning.

We refer to this failure mode as \emph{reward-profile collapse}. 
It begins with \emph{profile aliasing}: static aggregation can map qualitatively different reward profiles to the same scalar value. 
For example, under equal weights, \([1,0,0]\) and \([0,1,0]\) become indistinguishable after scalarization. 
This aliasing can then lead to \emph{shortcut lock-in}~\citep{geirhos2020shortcut}: easy or dense dimensions dominate early scalar reward improvements, while harder dimensions remain under-optimized. 
In mathematical reasoning, as shown in Figure~\ref{fig:intro-collapse}, format and length rewards provide faster and denser feedback than answer correctness, so the policy first learns to satisfy superficial constraints while still failing the task. Once such shortcut profiles dominate rollouts, the policy receives fewer informative samples for neglected dimensions, trapping it in a local optimum under the scalarized objective, making later recovery difficult and ultimately preventing the policy from reaching better-balanced profiles that would yield higher task performance. Unlike conventional reward hacking~\citep{skalse2022defining,gao2023scaling}, this failure does not require any reward dimension to be invalid—even when all dimensions are meaningful, static aggregation alone can create the trap. 
We call this phenomenon \emph{aggregation-induced reward hacking}.

To tackle this issue, we propose \emph{Adaptive Multi-Reward Projection} (AMRP), a simple but effective aggregation method to mitigate this reward-profile collapse. 
Rather than projecting reward profiles with a fixed weight vector throughout training, AMRP adapts the projection online according to the observed dynamics of each reward dimension. 
Specifically, we introduce three signals: relative shortfall, reward volatility, and recent progress. 
Relative shortfall measures whether a dimension is lagging behind the others. 
Reward volatility measures whether a dimension remains unstable. 
Recent progress measures whether a dimension is improving, stagnant, or regressing. 
AMRP increases the weights of dimensions that are lagging, unstable, or stagnant, and decreases the weights of dimensions that are already saturated and stable. 
In this way, our method effectively mitigates the reward-profile collapse.
We evaluate AMRP across three different multi-reward settings: rule-based mathematical reasoning, citation-grounded generation with automatic evaluators, and open-ended alignment with learned reward-model scores. AMRP consistently improves downstream performance and reward-profile dynamics, with particularly large gains in mathematical reasoning and citation-grounded generation and consistent gains in the more correlated open-ended setting. We further show that AMRP composes effectively with GDPO and PPO and remains robust
across the tested hyperparameter ranges.

Our contributions can be summarized as: 
\begin{itemize}
\item We identify \emph{reward-profile collapse} as a mode of aggregation-induced reward hacking in multi-reward reinforcement learning, arising from profile aliasing and shortcut lock-in.
\item We propose \emph{Adaptive Multi-Reward Projection}, a dynamic aggregation method that adapts reward projection online according to three signals: relative shortfall, reward volatility, and recent progress.
\item We validate AMRP across structured reasoning, citation-grounded generation, and open-ended alignment with multiple backbones, showing consistent downstream gains and compatibility with GRPO, GDPO, and PPO.
\end{itemize}

\begin{figure*}[t]
\centering
\includegraphics[width=0.85\textwidth]{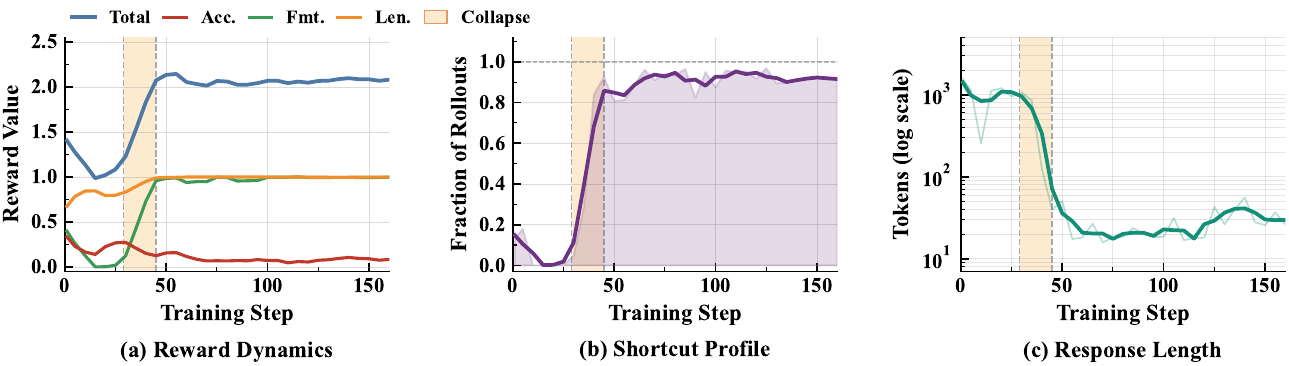}
\caption{
Reward-profile collapse under static weighting in math reasoning.
(a) The scalar reward increases while answer accuracy drops, driven by format and length rewards saturating.
(b) The shortcut profile \((r_{\mathrm{acc}}=0,r_{\mathrm{fmt}}=1,r_{\mathrm{len}}=1)\) rapidly becomes dominant.
(c) Mean response length drops sharply, consistent with a short format-compliant shortcut strategy.
}
\label{fig:math-static-collapse}
\end{figure*}

\section{Related Work}

\noindent\paragraph{Reward Hacking and Overoptimization.}
Reward hacking arises when optimizing an imperfect proxy reward degrades performance on the true objective~\citep{skalse2022defining}. It is closely related to shortcut learning, where models exploit spurious features rather than the desired solution~\citep{geirhos2020shortcut,yuan2024llms}. 
In RLHF, strong optimization of learned reward models can cause overoptimization or Goodharting~\citep{gao2023scaling,coste2023reward}. 
Existing mitigations include constrained optimization~\citep{moskovitz2024confronting,ackermann2026gradient}, reward ensembles~\citep{coste2023reward,rame2024warm}, debiased reward learning~\citep{chen2024odin,miao2024inform}, and reward transformation~\citep{fu2025reward}. 
Our work studies a complementary failure arising from aggregation itself: static scalarization can obscure dimension-wise signals and steer optimization toward easier or denser dimensions, even when each reward is meaningful.

\noindent\paragraph{Multi-Dimensional Rewards.}
Modern LLM post-training combines heterogeneous reward sources. 
RLVR uses deterministic or programmatic feedback such as answer verification, code execution, and format constraints~\citep{shao2024deepseekmath,le2022coderl}. 
Grounded generation uses automatic evaluators for correctness, citation quality, and fluency, as in ALCE~\citep{gao2023enabling}. 
Open-ended alignment often relies on multi-attribute reward data or learned multi-objective reward models such as HelpSteer2~\citep{wang2024helpsteer} and ArmoRM~\citep{wang2024interpretable}. 
LLM-as-a-judge evaluation enables scalable assessment but can introduce biases such as position or verbosity preferences~\citep{zheng2023judging}. 
When multiple dimensions are jointly optimized in RL, they are typically scalarized before policy optimization.

\noindent\paragraph{Multi-Reward Aggregation.}
Since response quality is multidimensional, multi-reward optimization commonly scalarizes rewards with a fixed weighted sum, i.e., a static projection from reward profiles to scalar signals. 
When dimensions differ in density, variance, scale, or difficulty, this projection can discard learning-relevant information. 
DRBO dynamically reweights rewards~\citep{chen-etal-2025-drbo} using single-statistic rules based on reward levels or recent reward changes. GDPO instead improves GRPO signal resolution through dimension-wise reward normalization~\citep{liu2026gdpo}. 
Our work targets aggregation-induced reward hacking, where static aggregation aliases reward profiles and induces shortcut lock-in even with valid rewards. AMRP combines shortfall, volatility, and progress signals as a soft conjunction, with EMA smoothing for stability. It therefore differs from DRBO-style single-statistic reweighting and complements GDPO-style normalization.

\section{Aggregation-Induced Reward Hacking}
\label{sec:hacking}

This section formalizes how reward aggregation itself can induce reward hacking. 
We view static aggregation as a projection from a multi-dimensional reward profile to a scalar training signal. Although computationally simple, this projection discards dimension-wise information, leading to \emph{profile aliasing}, where distinct profiles become indistinguishable, and \emph{shortcut lock-in}, where optimization exploits easy dimensions while weakening signals from harder ones. We then provide diagnostic evidence from structured reasoning experiments.

\subsection{Static Aggregation as Projection}

We consider multi-reward reinforcement learning for LLM post-training. 
Given a prompt \(x\sim\mathcal{D}\), a policy \(\pi_\theta\) generates a response \(y\sim\pi_\theta(\cdot|x)\), which is evaluated by \(K\) reward dimensions:
\begin{equation}
\mathbf{r}(x,y)=[r_1(x,y),\ldots,r_K(x,y)].
\end{equation}
Each dimension measures one aspect of response quality (e.g., correctness, format or length).
A common practice is to aggregate the reward profile into a scalar reward:
\begin{equation}
R_{\mathbf{w}}(x,y)=\sum_{i=1}^{K} w_i r_i(x,y),
\end{equation}
where weight \(w_i\ge 0\) and \(\frac{1}{K}\sum_i w_i=1\). This aggregation maps the \(K\)-dimensional reward profile to a one-dimensional signal. Once \(R_{\mathbf{w}}\) is used in the RL training objective, the policy update is driven by the scalarized reward rather than by the full reward profile, and distinctions among reward dimensions are no longer directly visible to the learning algorithm.
For example, in GRPO, a group of rollouts \(\{y_j\}_{j=1}^{G}\) from the same prompt is converted into group-relative advantages:
\begin{equation}
\hat{A}_j
=
\frac{
R_{\mathbf{w}}(x,y_j)-\operatorname{mean}_{k}R_{\mathbf{w}}(x,y_k)
}{
\operatorname{std}_{k}R_{\mathbf{w}}(x,y_k)+\epsilon
}.
\end{equation}
Therefore, the projection weights \(\mathbf{w}\) determine the scalar reward
gaps within each rollout group, which in turn determine the relative advantages used for policy updates. As a result, different reward profiles that induce the same scalarized values are treated identically during training, even if they reflect qualitatively different behaviors.

\subsection{Reward-Profile Collapse}
We define \emph{reward-profile collapse} as the loss of dimension-wise learning information induced by static reward aggregation. 
Because only scalarized rewards are visible during policy updates, distinct profiles can become indistinguishable,
biasing learning toward easier dimensions.

\paragraph{Profile Aliasing.}
Static aggregation can map different reward profiles to the same or similar
scalar value.
For example, under static weights \{1,1,1\}, different reward profiles $[1,0,0],\ [0,1,0],\ [0,0,1]$ are projected into the same scalar reward, although they correspond to very different cases.
In mathematical reasoning, these could correspond to a correct but poorly formatted answer, a formatted but incorrect answer, and a response that only satisfies a length constraint. 
After scalarization, dimension-wise differences are invisible to the policy update.

\paragraph{Shortcut Lock-in.}
This loss can compound during training. 
Early in training, easy or dense dimensions can improve faster than harder or sparser dimensions. 
For example, a policy may learn format and length compliance before answer correctness, moving toward shortcut profiles that achieve high scalar reward without solving the problem. 
Once these profiles dominate rollouts, harder dimensions may become sparse or low-contrast within rollout groups, making it difficult for group-relative advantages to reinforce the desired behavior.

Together, profile aliasing and shortcut lock-in lead to \emph{aggregation-induced reward hacking}: the policy can increase the scalar reward by exploiting easy, dense, or reward-model-preferred dimensions rather than improving the full reward profile. 
Such shortcut profiles can become locally attractive under the scalarized objective and dominate rollouts, leaving neglected dimensions with fewer informative samples and weaker learning signals, which makes recovery difficult and prevents convergence to balanced profiles with higher task performance.

\begin{algorithm}[t]
\small
\caption{AMRP}
\label{alg:amrp}
\KwIn{Reward functions $\{r_i\}_{i=1}^{K}$; initial weights $\mathbf w^{(0)}$;
projection prior $\mathbf a$; update interval $N$; EMA decay $\rho$}
\KwOut{Scalar reward $R^{(t)}(x,y)$}

\For{training step $t=1,2,\ldots$}{
    Generate rollouts and compute reward profile $\mathbf r(x,y)$\;

    Compute $R^{(t)}(x,y)\leftarrow
    \sum_{i=1}^{K}w_i^{(t-1)}r_i(x,y)$\;
    Update policy with scalar reward $R^{(t)}$\;

    Update or initialize EMA statistics
    $\mu_i^{(t)},\nu_i^{(t)},\sigma_i^{(t)},\Delta_i^{(t)}$\;

    \If{$t \bmod N = 0$}{
        Compute gates $S_i^{(t)},V_i^{(t)},P_i^{(t)}$ for all $i$\;
        Set $U_i^{(t)}\leftarrow
        S_i^{(t)}V_i^{(t)}P_i^{(t)}$\;
        Set $w_i^{(t)}\leftarrow
        K a_iU_i^{(t)}
        /\sum_{j=1}^{K}a_jU_j^{(t)}$\;
    }
    \Else{
        Set $w_i^{(t)}\leftarrow w_i^{(t-1)}$\;
    }
}
\end{algorithm}

\subsection{Diagnostic Evidence}

We use mathematical reasoning as a diagnostic case with discrete, interpretable rewards: $\mathbf{r}_{\mathrm{math}}=[r_{\mathrm{acc}},r_{\mathrm{fmt}},r_{\mathrm{len}}]$,
measuring answer correctness, format compliance, and length control.
Figure~\ref{fig:math-static-collapse} shows reward-profile collapse under static equal weighting.
Between steps 35 and 40, scalar reward rises from 1.445 to 1.973, while accuracy drops from 0.258 to 0.090.
The gain is driven by format and length rewards, which rise to near saturation, revealing the shortcut profile \([0,1,1]\): an incorrect response satisfying auxiliary constraints.
Although inferior to \([1,1,1]\), it receives high scalar reward and quickly dominates rollouts, accompanied by collapsed response lengths.

Early rollouts follow the pattern in Figure~\ref{fig:intro-collapse}: short, format-compliant responses satisfy \(r_{\mathrm{fmt}}\) and \(r_{\mathrm{len}}\) but fail \(r_{\mathrm{acc}}\). 
As such samples become common, rollout groups contain few correct responses, weakening group-relative signals for answer correctness, making later recovery difficult.
This motivates adaptive aggregation that down-weights saturated rewards and shifts pressure toward lagging, unstable, or stagnant dimensions.

\section{Adaptive Multi-Reward Projection}
\label{sec:method}

Section~\ref{sec:hacking} shows that static aggregation can fail because it uses a static projection throughout training. 
We mitigate this by making the projection adaptive. 
Instead of scalarizing reward profiles with a fixed vector \(\mathbf w\), AMRP adapts the vector \(\mathbf w^{(t)}\) at each step $t$ according to online reward dynamics:
{
\setlength{\belowdisplayskip}{3pt}
\begin{equation}
R^{(t)}(x,y)=\sum_{i=1}^{K}w_i^{(t-1)}r_i(x,y).
\label{eq:amrp-reward}
\end{equation}
}

Algorithm~\ref{alg:amrp} summarizes the AMRP update procedure. The goal is to increase the influence of reward dimensions that are lagging, unstable, or stagnant, while reducing pressure on dimensions that are already saturated. 
We assume all reward dimensions are oriented so that larger values are better and are normalized to comparable scales.

\subsection{Adaptive Projection Weighting}
AMRP adapts projection weights using three online signals designed to address the failure modes of reward-profile collapse described in Section~\ref{sec:hacking}.
1) \textit{Relative shortfall} prevents low-performing dimensions from being hidden by high-scoring rewards. 
2) \textit{Reward volatility} keeps pressure on dimensions that remain unstable. 
3) \textit{Recent progress} detects stagnation or regression, which may indicate that a dimension is losing optimization pressure.

Let \(\mu_i^{(t)}\), \(\sigma_i^{(t)}\), and \(\Delta_i^{(t)}\) denote the estimated mean, variability, and recent progress of reward dimension \(i\), and let
\(\bar{\mu}^{(t)}=\frac{1}{K}\sum_{j=1}^{K}\mu_j^{(t)}\). 
AMRP computes three gates: 
{
\setlength{\belowdisplayskip}{3pt}
\begin{align}
S_i^{(t)}
&=
\operatorname{softplus}\!\left(k_s(\bar{\mu}^{(t)}-\mu_i^{(t)})\right), \nonumber\\
V_i^{(t)}
&=
1+\lambda\frac{\sigma_i^{(t)}}{\mu_i^{(t)}+\epsilon}, \label{eq:amrp-gates}\\
P_i^{(t)}
&=
\operatorname{softplus}\!\left(-k_p\Delta_i^{(t)}\right), \nonumber
\end{align}}
which emphasize dimensions that are lagging, unstable, or stagnant, respectively.
They are then combined into a priority score and normalized into mean-one projection weights:
{
\setlength{\belowdisplayskip}{3pt}
\begin{align}
U_i^{(t)}
&=
S_i^{(t)}V_i^{(t)}P_i^{(t)}, \nonumber\\
w_i^{(t)}
&=
K\cdot
\frac{U_i^{(t)}}{\sum_{j=1}^{K}U_j^{(t)}} .
\label{eq:amrp-weight}
\end{align}}

Thus, \(w_i^{(t)}>1\) amplifies a dimension relative to static equal weighting, while \(w_i^{(t)}<1\) reduces its relative influence. 
The multiplicative score acts as a soft conjunction: a dimension receives high priority when multiple signals indicate under-optimization.

\paragraph{Non-uniform projection priors.}
AMRP can incorporate a mean-one prior vector \(\mathbf a\) by replacing \(U_i^{(t)}\) with \(a_iU_i^{(t)}\) in Eq.~\ref{eq:amrp-weight}. 
This changes the preferred projection direction while retaining online adaptation. 
The default setting is \(a_i=1\) for all dimensions.

\newcommand{\drboinv}{\textit{DRBO}\textsubscript{\textit{inverse}}}
\newcommand{\drbodelta}{\textit{DRBO}\textsubscript{\textit{delta}}}

\begin{table*}[t]
\centering
\footnotesize
\setlength{\tabcolsep}{5pt}
\renewcommand{\arraystretch}{1.12}

\makebox[\textwidth][c]{%
\begin{tabular}{
@{}l@{\hspace{10pt}}
ccc@{\hspace{14pt}}
ccc@{\hspace{14pt}}
ccc@{\hspace{14pt}}
cccc@{}
}
\toprule

\multirow{2}{*}{\textbf{Method}}
& \multicolumn{3}{c}{\textit{\textbf{Math}}}
& \multicolumn{3}{c}{\textit{\textbf{AMC}}}
& \multicolumn{3}{c}{\textit{\textbf{AIME}}}
& \multicolumn{4}{c}{\textbf{Average}} \\

\cmidrule(l{2pt}r{18pt}){2-4}
\cmidrule(l{2pt}r{18pt}){5-7}
\cmidrule(l{2pt}r{18pt}){8-10}
\cmidrule(l{2pt}r{2pt}){11-14}

& \textbf{Acc.} & \textbf{Fmt.} & \textbf{Len.}
& \textbf{Acc.} & \textbf{Fmt.} & \textbf{Len.}
& \textbf{Acc.} & \textbf{Fmt.} & \textbf{Len.}
& \textbf{Acc.} & \textbf{Fmt.} & \textbf{Len.}
& \textbf{Overall} \\

\specialrule{\lightrulewidth}{1pt}{1pt}

\rowcolor{gray!15}
\multicolumn{14}{c}{\textbf{Qwen3-4B-Instruct}} \\

Base
& 86.40 & 47.60 & 86.52
& 55.31 & 20.18 & 56.90
& 26.04 & 8.96 & 25.83
& 55.92 & 25.58 & 56.42 & 45.97 \\

Static
& 72.00 & 96.60 & 96.48
& 33.81 & 93.03 & 92.62
& 6.77 & 93.44 & 93.44
& 37.53 & 94.36 & 94.18 & 75.35 \\

\drbodelta
& 81.60 & \textbf{97.80} & \textbf{97.67}
& 53.92 & \textbf{97.63} & \textbf{97.46}
& 17.92 & 91.35 & 91.32
& 51.15 & 95.59 & 95.48 & \textbf{80.74} \\

\drboinv
& 78.00 & 97.60 & 97.35
& 47.25 & 96.31 & 96.19
& 5.31 & \textbf{97.19} & \textbf{97.19}
& 43.52 & \textbf{97.03} & \textbf{96.91} & 79.15 \\

\cdashline{1-14}
\noalign{\vskip 2pt}

AMRP
& \textbf{86.60} & 93.20 & 92.58
& \textbf{63.59} & 74.66 & 73.67
& \textbf{31.98} & 42.71 & 41.77
& \textbf{60.72} & 70.19 & 69.34 & 66.75 \\

\specialrule{\lightrulewidth}{1pt}{1pt}

\rowcolor{gray!15}
\multicolumn{14}{c}{\textbf{DeepSeek-Math-7B-Base}} \\

Base
& 21.00 & 33.40 & 75.59
& 3.84 & 16.30 & 69.20
& 0.62 & 10.10 & 60.73
& 8.49 & 19.93 & 68.51 & 32.31 \\

Static
& 36.00 & 97.40 & 97.40
& 11.60 & \textbf{99.02} & \textbf{99.02}
& 0.00 & \textbf{96.98} & \textbf{96.98}
& 15.87 & \textbf{97.80} & \textbf{97.80} & 70.49 \\

\drbodelta
& 34.00 & 81.60 & 54.80
& 14.53 & 70.63 & 55.53
& 0.62 & 43.65 & 26.67
& 16.38 & 65.29 & 45.67 & 42.45 \\

\drboinv
& 35.40 & 95.80 & 96.60
& 11.90 & 96.12 & 97.14
& 0.73 & 87.60 & 89.06
& 16.01 & 93.17 & 94.27 & 67.82 \\

\cdashline{1-14}
\noalign{\vskip 2pt}

AMRP
& \textbf{38.00} & \textbf{99.00} & \textbf{99.00}
& \textbf{14.57} & \textbf{99.02} & \textbf{99.02}
& \textbf{3.23} & 93.54 & 93.33
& \textbf{18.60} & 97.19 & 97.12 & \textbf{70.97} \\

\bottomrule
\end{tabular}%
}

\caption{
Performance on MATH, AMC, and AIME.
Acc., Fmt., and Len. denote accuracy, format compliance, and length-control scores, respectively, all on a 0--100 scale.
The Average block reports the macro-average of each criterion across the three benchmarks;
Overall is the arithmetic mean of the three averaged criteria.
}
\label{tab:math}
\end{table*}

\subsection{Online Estimation and Weight Updates}

AMRP estimates reward statistics from training batches. 
Let \(\bar r_{i,B}^{(t)}\) and \(v_{i,B}^{(t)}\) denote the batch mean and variance of reward dimension \(i\).
We initialize \(\mu_i^{(1)}=\bar r_{i,B}^{(1)}\), \(\nu_i^{(1)}=v_{i,B}^{(1)}\), and \(\Delta_i^{(1)}=0\).
For \(t>1\), we maintain EMA estimates:
{
\setlength{\belowdisplayskip}{5pt}
\begin{align}
\mu_i^{(t)}
&= \rho\mu_i^{(t-1)}
 + (1-\rho)\bar r_{i,B}^{(t)}, \nonumber\\
\nu_i^{(t)}
&= \rho\nu_i^{(t-1)}
 + (1-\rho)v_{i,B}^{(t)}, \nonumber\\
\sigma_i^{(t)}
&= \sqrt{\nu_i^{(t)}}, 
\qquad
\Delta_i^{(t)}
= \mu_i^{(t)}-\mu_i^{(t-1)} .
\label{eq:amrp-ema}
\end{align}}

Projection weights are refreshed every \(N\) training steps and reused between refreshes, which reduces sensitivity to batch-level noise.

\subsection{Interface with Policy Optimization}

AMRP operates at the cross-reward aggregation interface and is therefore compatible with different policy optimization methods.
For scalar-reward optimizers such as GRPO and PPO, AMRP adaptively scalarizes the reward profile before advantage estimation; in GRPO, the weights $w_i^{(t)}$ directly shape reward gaps within rollout groups and hence the group-relative advantages.
For dimension-wise methods such as GDPO, the same online weights can aggregate normalized per-reward advantages before final advantage normalization.
Thus, AMRP provides adaptive cross-reward prioritization without modifying the underlying policy objective.

\begin{table*}[t]
\centering
\footnotesize
\setlength{\tabcolsep}{12pt}
\renewcommand{\arraystretch}{1.12}

\makebox[\textwidth][c]{%
\begin{tabular}{
@{}l@{\hspace{20pt}}
ccc@{\hspace{28pt}}
ccc@{\hspace{28pt}}
ccc@{}
}
\toprule

\multirow{2}{*}{\textbf{Method}}
& \multicolumn{3}{c}{\textit{\textbf{ASQA}}}
& \multicolumn{3}{c}{\textit{\textbf{ELI5}}}
& \multicolumn{3}{c}{\textbf{Average}} \\

\cmidrule(l{2pt}r{30pt}){2-4}
\cmidrule(l{2pt}r{30pt}){5-7}
\cmidrule(l{2pt}r{2pt}){8-10}

& \textbf{Cor.} & \textbf{Cit.} & \textbf{Avg.}
& \textbf{Cor.} & \textbf{Cit.} & \textbf{Avg.}
& \textbf{Cor.} & \textbf{Cit.} & \textbf{Overall} \\

\specialrule{\lightrulewidth}{1pt}{1pt}

\rowcolor{gray!15}
\multicolumn{10}{c}{\textbf{Qwen3-4B-Instruct}} \\

Base
& 52.10 & 57.89 & 54.99
& 22.17 & 38.99 & 30.58
& 37.14 & 48.44 & 42.79 \\

Static
& 58.75 & 82.76 & 70.75
& 18.00 & 44.62 & 31.31
& 38.38 & 63.69 & 51.03 \\

\drbodelta
& 60.83 & 59.74 & 60.28
& \textbf{24.67} & 33.96 & 29.32
& 42.75 & 46.85 & 44.80 \\

\drboinv
& 58.86 & \textbf{89.75} & 74.31
& 21.17 & \textbf{60.54} & 40.85
& 40.02 & \textbf{75.15} & 57.58 \\

\cdashline{1-10}
\noalign{\vskip 2pt}

AMRP
& \textbf{63.90} & 89.60 & \textbf{76.75}
& 23.50 & 59.51 & \textbf{41.51}
& \textbf{43.70} & 74.56 & \textbf{59.13} \\

\specialrule{\lightrulewidth}{1pt}{1pt}

\rowcolor{gray!15}
\multicolumn{10}{c}{\textbf{Llama-3.1-8B-Instruct}} \\

Base
& 45.80 & 54.74 & 50.27
& 21.67 & 38.87 & 30.27
& 33.74 & 46.81 & 40.27 \\

Static
& 60.01 & 90.29 & 75.15
& 14.83 & 71.71 & 43.27
& 37.42 & 81.00 & 59.21 \\

\drbodelta
& 63.50 & 73.33 & 68.41
& 21.33 & 33.34 & 27.34
& 42.42 & 53.34 & 47.88 \\

\drboinv
& 56.47 & 88.11 & 72.29
& 49.67 & 70.14 & 59.91
& 53.07 & 79.13 & 66.10 \\

\cdashline{1-10}
\noalign{\vskip 2pt}

AMRP
& \textbf{66.10} & \textbf{93.80} & \textbf{79.95}
& \textbf{65.33} & \textbf{76.98} & \textbf{71.16}
& \textbf{65.72} & \textbf{85.39} & \textbf{75.56} \\

\bottomrule
\end{tabular}%
}

\caption{
Performance on ASQA and ELI5.
Each benchmark reports correctness (Cor.), citation (Cit.), and their average (Avg.), all on a 0–100 scale.
The Average block reports the macro-average of each criterion across the two benchmarks;
Overall is the arithmetic mean of the averaged correctness and citation scores.
}
\label{tab:asqa-eli5}
\end{table*}

\begin{table*}[t]
\centering
\footnotesize
\setlength{\tabcolsep}{5pt}
\renewcommand{\arraystretch}{1.12}

\makebox[\textwidth][c]{%
\begin{tabular}{
@{}l@{\hspace{6pt}}
ccc@{\hspace{8pt}}
ccc@{\hspace{8pt}}
ccc@{\hspace{8pt}}
cccc@{}
}
\toprule

\multirow{2}{*}{\textbf{Method}}
& \multicolumn{3}{c}{\textit{\textbf{AlpacaEval}}}
& \multicolumn{3}{c}{\textit{\textbf{ArenaHard}}}
& \multicolumn{3}{c}{\textit{\textbf{MT-Bench}}}
& \multicolumn{4}{c}{\textbf{Average}} \\

\cmidrule(l{2pt}r{18pt}){2-4}
\cmidrule(l{2pt}r{18pt}){5-7}
\cmidrule(l{2pt}r{18pt}){8-10}
\cmidrule(l{2pt}r{2pt}){11-14}

& \textbf{Help.} & \textbf{Corr.} & \textbf{Coher.}
& \textbf{Help.} & \textbf{Corr.} & \textbf{Coher.}
& \textbf{Help.} & \textbf{Corr.} & \textbf{Coher.}
& \textbf{Help.} & \textbf{Corr.} & \textbf{Coher.}
& \textbf{Overall} \\

\specialrule{\lightrulewidth}{1pt}{1pt}

\rowcolor{gray!15}
\multicolumn{14}{c}{\textbf{Qwen3-4B-Instruct}} \\

Base
& 55.95 & 55.63 & 64.48
& 49.89 & 50.73 & 58.69
& 59.46 & 59.38 & 66.94
& 55.10 & 55.25 & 63.37 & 57.91 \\

Static
& 83.10 & 81.57 & 85.34
& 71.25 & 71.00 & \textbf{73.20}
& 81.16 & 79.12 & 84.52
& 78.50 & 77.23 & 81.02 & 78.92 \\

\drbodelta
& 83.38 & 81.79 & 85.55
& 71.42 & 71.14 & \textbf{73.20}
& 80.73 & 78.81 & 84.43
& 78.51 & 77.25 & 81.06 & 78.94 \\

\drboinv
& 83.35 & 81.91 & 85.76
& 70.88 & 70.62 & 72.78
& 80.24 & 78.55 & 84.60
& 78.16 & 77.03 & 81.05 & 78.74 \\

\cdashline{1-14}
\noalign{\vskip 2pt}

AMRP
& \textbf{84.06} & \textbf{82.50} & \textbf{86.10}
& \textbf{71.69} & \textbf{71.57} & \textbf{73.20}
& \textbf{81.38} & \textbf{79.74} & \textbf{85.27}
& \textbf{79.04} & \textbf{77.94} & \textbf{81.52}
& \textbf{79.50} \\

\specialrule{\lightrulewidth}{1pt}{1pt}

\rowcolor{gray!15}
\multicolumn{14}{c}{\textbf{Llama-3.1-8B-Instruct}} \\

Base
& 80.60 & 79.55 & 82.43
& 71.05 & 70.87 & 72.66
& 80.38 & 78.84 & 83.39
& 77.34 & 76.42 & 79.49 & 77.75 \\

Static
& 84.69 & 83.36 & 86.79
& 74.51 & 73.60 & 77.82
& 81.93 & 80.12 & \textbf{86.23}
& 80.38 & 79.03 & 83.61 & 81.01 \\

\drbodelta
& 84.55 & 83.31 & 86.56
& 74.27 & 73.37 & 77.51
& 82.39 & 80.63 & 86.10
& 80.40 & 79.10 & 83.39 & 80.97 \\

\drboinv
& 84.12 & 82.78 & 86.17
& 74.34 & 73.50 & 77.64
& 82.12 & 80.35 & 86.13
& 80.19 & 78.88 & 83.31 & 80.79 \\

\cdashline{1-14}
\noalign{\vskip 2pt}

AMRP
& \textbf{85.12} & \textbf{83.83} & \textbf{86.95}
& \textbf{75.25} & \textbf{74.31} & \textbf{78.53}
& \textbf{82.85} & \textbf{81.07} & 86.04
& \textbf{81.07} & \textbf{79.74} & \textbf{83.84}
& \textbf{81.55} \\

\bottomrule
\end{tabular}%
}

\caption{
ArmoRM-scored evaluation on held-out AlpacaEval, ArenaHard, and MT-Bench prompts after HelpSteer2 training.
Scores are scaled to $[0,100]$.
Help., Corr., and Coher.\ denote helpfulness, correctness, and coherence.
Average reports criterion-wise macro-averages across prompt sets, and Overall their arithmetic mean.
}
\label{tab:helpsteer}
\end{table*}

\section{Experiments}
\subsection{Experimental Setups}
\label{sec:exp-setup}

\paragraph{Training Protocol.}
We use full-parameter GRPO fine-tuning~\citep{shao2024deepseekmath} with group size \(G=8\) across all three settings.
Methods share the same pipeline and differ only in projection weighting.
Rewards are scaled to \([0,1]\), and training rollouts use a sampling temperature of \(1.0\).
By default, AMRP initializes \(w_i^{(0)}=a_i=1\) for all reward dimensions and updates the projection weights every \(N\) steps.
Experiments run on NVIDIA A100 GPUs; see Appendix~\ref{app:implementation} for details.

\paragraph{Structured Reasoning.}
We study mathematical reasoning with verifiable rewards.
Following DeepSeek-style rule-based RLVR with accuracy and format rewards~\citep{guo2025deepseek}, we additionally include a soft length-control reward to model the generation budget.
We train on a 10K-sample subset of NuminaMath-1.5~\citep{numina_math_datasets} and evaluate on MATH-500~\citep{lightman2024let}, AMC\footnote{\url{https://huggingface.co/datasets/AI-MO/aimo-validation-amc}}, and AIME\footnote{\url{https://huggingface.co/datasets/AI-MO/aimo-validation-aime}}.
The reward profile is
\(\mathbf r_{\mathrm{math}}=[r_{\mathrm{acc}},r_{\mathrm{fmt}},r_{\mathrm{len}}]\),
covering final-answer accuracy, format compliance, and length control.
We fine-tune Qwen3-4B-Instruct~\citep{yang2025qwen3} and DeepSeek-Math-7B-Base~\citep{shao2024deepseekmath} for one epoch with a maximum completion length of 3,000 tokens.

\paragraph{Grounded Generation.}
We study citation-grounded long-form QA following ALCE~\citep{gao2023enabling} and the setup of DRBO~\citep{chen-etal-2025-drbo} on ASQA~\citep{stelmakh2022asqa} and ELI5~\citep{fan2019eli5}.
Each query is provided with three retrieved passages and one in-context demonstration.
The reward profile is
\(\mathbf r_{\mathrm{alce}}=[r_{\mathrm{corr}},r_{\mathrm{cite}}]\),
measuring task-specific correctness and citation quality.
We train Qwen3-4B-Instruct~\citep{yang2025qwen3} and Llama-3.1-8B-Instruct~\citep{grattafiori2024llama} for three epochs, using maximum completion lengths of 400 and 500 tokens for ASQA and ELI5, respectively.

\paragraph{Open-ended Alignment.}
We train on HelpSteer2~\citep{wang2024helpsteer} and score responses online using ArmoRM-Llama3-8B-v0.1~\citep{wang2024interpretable}. 
The reward profile is \(\mathbf r_{\mathrm{align}}=[r_{\mathrm{help}},r_{\mathrm{corr}},r_{\mathrm{coh}}]\), corresponding to helpfulness, correctness, and coherence. 
We use Qwen3-4B-Instruct~\citep{yang2025qwen3} and Llama-3.1-8B-Instruct~\citep{grattafiori2024llama}, and train for 3 epochs with a maximum completion length of 512. 
We evaluate on held-out prompts from AlpacaEval~\citep{dubois2024length}, ArenaHard~\citep{li2024crowdsourced}, and MT-Bench~\citep{zheng2023judging}, reporting ArmoRM-head scores scaled to \([0,100]\).

\paragraph{Baselines.}
We compare AMRP with static equal weighting and two dynamic weighting baselines~\citep{chen-etal-2025-drbo}: DRBO\textsubscript{\textit{inverse}}, which up-weights lower-scoring dimensions, and DRBO\textsubscript{\textit{delta}}, which up-weights dimensions showing stronger recent improvement. We report the untrained base model before GRPO fine-tuning as \textit{Base}.

\begin{figure}[t]
\centering
\includegraphics[width=\linewidth]{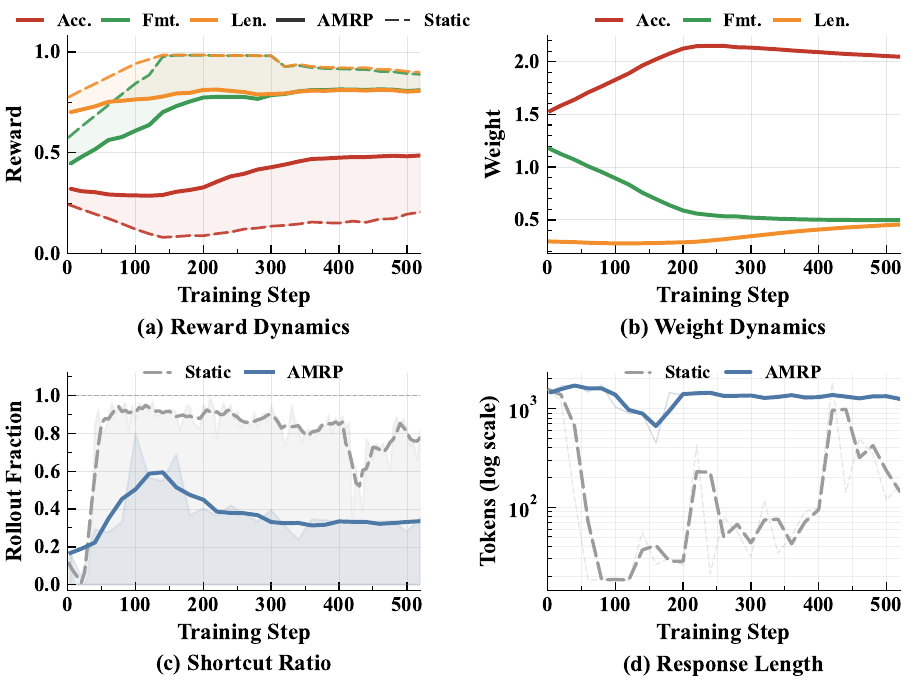}
\caption{
Mechanistic analysis in math reasoning with Qwen3-4B-Instruct.
(a,b) AMRP improves accuracy by up-weighting accuracy and down-weighting saturated format and length rewards.
(c,d) Static aggregation induces shortcut profiles \((r_{\mathrm{acc}},r_{\mathrm{fmt}},r_{\mathrm{len}})=(0,1,1)\) and collapsed response lengths, whereas AMRP reduces shortcuts and maintains longer generations.
}
\label{fig:mechanism}
\end{figure}

\subsection{Main Results}
\label{sec:main-results}

\paragraph{Structured Reasoning.}
In mathematical reasoning, static aggregation often over-optimizes format and length rewards while leaving answer accuracy under-optimized.
As shown in Table~\ref{tab:math}, AMRP achieves the best accuracy on all three benchmarks for both base models, with particularly large gains on the more challenging AMC and AIME benchmarks.
For Qwen3-4B-Instruct, AMRP improves AIME accuracy from 17.92 to 31.98 and AMC accuracy from 53.92 to 63.59 over the strongest baseline.
Format and length scores are more sensitive to dataset difficulty: harder problems often require longer reasoning trajectories, making responses more likely to reach the generation limit before completing the reasoning and closing the required tags, which reduces both format and length rewards.
AMRP therefore prioritizes correctness when additional reasoning is needed, rather than favoring short, format-compliant shortcuts.

\paragraph{Grounded Generation.}
In citation-grounded QA, static aggregation can induce aggregation-driven reward hacking, where citation support is over-optimized while correctness remains under-optimized.
As shown in Table~\ref{tab:asqa-eli5}, AMRP substantially improves correctness while maintaining strong citation quality, yielding the best overall trade-off for both models.
The overall score increases from 51.03 to 59.13 on Qwen3-4B-Instruct and from 59.21 to 75.56 on Llama-3.1-8B-Instruct.

\paragraph{Open-ended Alignment.}
In open-ended alignment, the learned reward dimensions are continuous and highly correlated, indicating that helpfulness, correctness, and coherence largely improve together. This leaves less aggregation-induced imbalance for AMRP to correct and, together with the already strong static baseline, provides less headroom for further gains. Nevertheless, Table~\ref{tab:helpsteer} shows that AMRP consistently improves both backbones and achieves the best overall score across AlpacaEval, ArenaHard, and MT-Bench. These results suggest that adaptive projection remains effective even in a milder setting with continuous, highly correlated model-based rewards.

\paragraph{Summary.}
Across all three settings, AMRP achieves the strongest reward-profile trade-offs by reallocating pressure from saturated dimensions to low, unstable, or stagnant ones, mitigating shortcut-dominated optimization under static scalarization.

\subsection{Mechanistic Analysis}
\label{sec:mechanistic}

The main results show that AMRP improves final reward-profile trade-offs.
We further examine how AMRP achieves these gains.
Figure~\ref{fig:mechanism} analyzes this process in mathematical reasoning, where both reward dimensions and shortcut behaviors are directly interpretable.

\paragraph{Projection Dynamics.}
Figures~\ref{fig:mechanism} (a,b) show the projection-level effect. 
Under static aggregation, format and length rewards quickly approach saturation, while accuracy remains much lower, indicating that the scalar objective is dominated by easier auxiliary dimensions. 
AMRP changes this trajectory by increasing \(w_{\mathrm{acc}}\) when accuracy lags and reducing the relative weights of saturated format and length rewards. 
This suggests that the improvement is driven by online reallocation of optimization pressure according to reward dynamics, rather than by a fixed accuracy bias.

\begin{table}[t]
\centering
\resizebox{\columnwidth}{!}{
\begin{tabular}{lcccccc}
\toprule
\textbf{Method}
& \textbf{Math}
& \textbf{AMC}
& \textbf{AIME}
& \textbf{Acc.}
& \textbf{Fmt.}
& \textbf{Len.} \\
\specialrule{\lightrulewidth}{1pt}{1pt}
\rowcolor{gray!15}
\multicolumn{7}{c}{\textbf{GDPO}} \\

Static
& 84.60 & 53.01 & 20.00
& 52.54 & 86.20 & 85.83 \\
\drbodelta
& 81.80 & 50.53 & 21.77
& 51.37 & 23.66 & 55.01 \\
\drboinv
& 74.40 & 44.47 & 10.00
& 42.96 & \textbf{99.61} & \textbf{99.61} \\
AMRP
& \textbf{84.80} & \textbf{60.24} & \textbf{26.67}
& \textbf{57.24} & 66.38 & 65.79 \\

\specialrule{\lightrulewidth}{1pt}{1pt}

\rowcolor{gray!15}
\multicolumn{7}{c}{\textbf{PPO}} \\

Static
& 86.40 & 56.63 & 20.00
& 54.34 & \textbf{53.47} & 55.76 \\
\drbodelta
& 80.80 & 50.19 & 17.71
& 49.57 & 22.68 & 49.06 \\
\drboinv
& 85.80 & 59.75 & 21.46
& 55.67 & 2.40 & 58.78 \\
AMRP
& \textbf{87.40} & \textbf{62.65} & \textbf{23.33}
& \textbf{57.79} & 49.33 & \textbf{59.42} \\

\bottomrule
\end{tabular}
}
\caption{
Compatibility of AMRP with GDPO and PPO on Qwen3-4B-Instruct mathematical reasoning.
Math, AMC, and AIME report accuracy;
Acc., Fmt., and Len.\ are macro-averages over the three benchmarks.
}
\label{tab:optimizer_compatibility}
\end{table}

\paragraph{Shortcut Suppression.}
Figures~\ref{fig:mechanism} (c,d) show the corresponding behavioral effect. 
We measure shortcut lock-in by the fraction of rollouts whose reward profile matches the shortcut pattern:
\[
\mathrm{SR}
=
\frac{1}{M}
\sum_{m=1}^{M}
\mathbb{I}\!\left[
(r_{\mathrm{acc}}^{(m)}, r_{\mathrm{fmt}}^{(m)}, r_{\mathrm{len}}^{(m)})
=
(0,1,1)
\right].
\]
Because this profile satisfies two of the three reward dimensions, static aggregation can assign it a high scalar reward even though it fails the task. 
Indeed, shortcut profiles become frequent under static aggregation and are accompanied by sharply shortened responses, consistent with short format-compliant generations rather than genuine problem solving. 
AMRP substantially reduces the shortcut ratio and maintains longer generations, showing that adaptive projection makes these shortcut responses less attractive during policy optimization.

\paragraph{Cross-setting Analysis.}
Similar patterns appear in grounded generation and open-ended alignment.
Appendix~\ref{app:additional_analysis} provides the corresponding reward and weight curves, as well as case studies: in grounded generation, AMRP counteracts the tendency of citation rewards to dominate correctness, while in open-ended alignment it adapts to smaller but persistent gaps among helpfulness, correctness, and coherence.
Overall, AMRP mitigates projection imbalance across settings.

\begin{table}[t]
\centering
\resizebox{\columnwidth}{!}{
\begin{tabular}{lcccccc}
\toprule
\textbf{Method}
& \textbf{Math}
& \textbf{AMC}
& \textbf{AIME}
& \textbf{Acc.}
& \textbf{Fmt.}
& \textbf{Len.} \\
\midrule
Static
& 80.20 & 45.37 & 12.19
& 45.92 & \textbf{97.17} & \textbf{97.06} \\
Init
& 84.80 & 58.09 & 29.27
& 57.39 & 78.13 & 76.88 \\
Prior
& \textbf{87.00} & \textbf{63.89} & \textbf{29.69}
& \textbf{60.19} & 69.83 & 68.74 \\
\bottomrule
\end{tabular}
}
\caption{
Non-uniform projection preferences in Qwen3-4B-Instruct mathematical reasoning.
Static, Init, and Prior apply \(1.5{:}0.75{:}0.75\) as a fixed projection, initialization, and persistent AMRP prior, respectively.
Math, AMC, and AIME report accuracy;
Acc., Fmt., and Len.\ are macro-averages over the three benchmarks.
}
\label{tab:nonuniform-math}
\end{table}

\subsection{Compatibility with RL Algorithms}
\label{sec:optimizer_compatibility}

AMRP operates at the reward-aggregation interface and is compatible with different policy optimizers. We evaluate it with GRPO, GDPO, and PPO on Qwen3-4B-Instruct mathematical reasoning. For GRPO and PPO, AMRP scalarizes rewards before advantage estimation; for GDPO, it aggregates normalized per-reward advantages, complementing GDPO's within-group normalization with adaptive cross-reward prioritization. For PPO, both the policy and value models are initialized from Qwen3-4B-Instruct. 
As shown in Table~\ref{tab:optimizer_compatibility}, AMRP consistently improves average accuracy across all three RL algorithms, from 37.53 to 60.72 with GRPO, 52.54 to 57.24 with GDPO, and 54.34 to 57.79 with PPO. GRPO with AMRP also substantially outperforms standard GDPO, while combining AMRP with GDPO provides further gains, supporting their complementarity. Under PPO, AMRP achieves the best average accuracy among the compared reward-aggregation methods. Overall, these results show that AMRP is not specific to GRPO and remains effective across different advantage-construction and policy-optimization schemes.

\subsection{Non-uniform Projection Priors}
\label{sec:nonuniform-prior}

Some tasks may prefer non-uniform reward priorities; for example, mathematical reasoning may emphasize accuracy while retaining format and length constraints.
We test an accuracy-biased ratio \(1.5{:}0.75{:}0.75\) over \([r_{\mathrm{acc}},r_{\mathrm{fmt}},r_{\mathrm{len}}]\) in three ways: as a fixed projection, as initialization only, and as the persistent AMRP prior introduced in Section~\ref{sec:method}.
Initialization-only uses the ratio only for \(\mathbf w^{(0)}\), after which weights follow the standard AMRP update rule; the persistent-prior variant keeps the ratio as the prior vector $\mathbf a$ throughout training.
As shown in Table~\ref{tab:nonuniform-math}, the fixed biased projection improves accuracy but remains limited.
Initialization-only AMRP further raises average accuracy to \(57.4\), showing that online adaptation can move beyond a suboptimal starting projection.
The persistent-prior variant performs best, reaching \(60.2\) average accuracy and the highest scores on Math, AMC, and AIME.
Thus, projection priors and adaptive weighting are complementary: priors encode task preference, while AMRP adjusts to reward dynamics and avoids over-optimizing auxiliary dimensions.

\begin{table}[t]
\centering
\small
\resizebox{\columnwidth}{!}{
\begin{tabular}{lcccccc}
\toprule
\textbf{Method}
& \textbf{Math}
& \textbf{AMC}
& \textbf{AIME}
& \textbf{Acc.}
& \textbf{Fmt.}
& \textbf{Len.} \\
\midrule
Full
& 86.60 & \textbf{63.59} & \textbf{31.98}
& \textbf{60.72} & 70.19 & 69.34 \\
w/o S
& 86.20 & 60.24 & 24.69
& 57.04 & 66.08 & 65.25 \\
w/o V
& 85.80 & 62.31 & 20.63
& 56.25 & 66.83 & 66.17 \\
w/o P
& 86.60 & 61.94 & 22.81
& 57.12 & 66.18 & 65.29 \\
Add.
& \textbf{87.20} & 61.67 & 23.96
& 57.61 & \textbf{71.77} & \textbf{71.08} \\
\bottomrule
\end{tabular}
}
\caption{
Ablation study on Qwen3-4B-Instruct mathematical reasoning.
S, V, and P denote the shortfall, volatility, and progress signals, respectively.
Add.\ replaces the multiplicative signal combination with an additive combination.
Math, AMC, and AIME report accuracy;
Acc., Fmt., and Len.\ are macro-averages over the three benchmarks.
}
\label{tab:ablation-math}
\end{table}

\subsection{Ablation Studies}
\label{sec:ablation}

We ablate AMRP by removing the shortfall, volatility, and progress signals one at a time, and by replacing the multiplicative rule with an additive variant.
Table~\ref{tab:ablation-math} shows that the full multiplicative design performs best overall, achieving the highest average accuracy as well as the best accuracy on AMC and AIME.
Removing any signal degrades performance, reducing average accuracy to \(57.0\), \(56.2\), and \(57.1\) without shortfall, volatility, and progress, respectively.
The largest drops occur on AIME, where removing volatility or progress lowers accuracy from \(31.9\) to \(20.6\) and \(22.8\).
Although the additive variant preserves slightly higher auxiliary rewards, it yields lower accuracy, supporting the multiplicative design as a soft conjunction that prioritizes dimensions only when multiple signals indicate under-optimization.

\subsection{Hyperparameter Sensitivity}
\label{sec:hyperparameter_sensitivity}

We evaluate AMRP's hyperparameter sensitivity on mathematical reasoning with Qwen3-4B-Instruct as a representative setting, varying one parameter at a time around the default $(\lambda,k_s,k_p,N)=(1.0,5,5,1)$. 
As shown in Table~\ref{tab:sensitivity_full} in the Appendix, average accuracy ranges from 54.45 to 60.72 across all tested configurations, consistently outperforming static aggregation at 37.53 and DRBO$_{\delta}$ at 51.15.
Each configuration also outperforms DRBO$_{\delta}$ on Math, AMC, and AIME, indicating that hyperparameters mainly affect gain magnitude rather than whether AMRP improves performance.
For the update interval $N\in\{1,2,4,8\}$, average accuracy remains stable between 58.33 and 60.72. Larger intervals remain competitive and do not degrade monotonically, while $N=1$ achieves the best overall accuracy, suggesting that AMRP is robust to the exact update frequency but benefits from more frequent updates.

\section{Conclusion}
\label{sec:conclusion}

In this paper, we identify \emph{reward-profile collapse} as a failure mode of multi-reward reinforcement learning, where static scalarization aliases distinct reward profiles and can lock optimization into easy, dense, or fast-improving dimensions.
This leads to \emph{aggregation-induced reward hacking}: scalar reward improves while important reward dimensions remain under-optimized.
To mitigate this problem, we propose \emph{Adaptive Multi-Reward Projection}, a lightweight aggregation method that adapts projection weights online using three signals: relative shortfall, reward volatility, and recent progress.
Across structured reasoning, grounded generation, and open-ended alignment, AMRP improves reward-profile trade-offs and downstream performance over static aggregation and DRBO-style dynamic weighting baselines.

\section*{Limitations}

This study evaluates AMRP across three representative multi-reward post-training settings involving rule-based rewards, automatic evaluators, and learned reward-model scores. While these settings cover diverse reward structures, they do not capture the full range of modern multi-reward pipelines. Our experiments also span multiple RL algorithms and reward dimensionalities, but broader evaluation with more numerous or strongly conflicting reward dimensions, larger-scale full-parameter training, and domains such as code generation, tool use, and safety-oriented alignment remains future work.

\section*{Ethical Considerations}

This work studies adaptive reward aggregation for multi-reward LLM post-training. AMRP mitigates reward imbalance but does not validate the rewards themselves; biases, factuality errors, verbosity preferences, or other artifacts in reward models and automatic evaluators may still propagate through optimization. Adaptive weighting is therefore not a substitute for careful reward design, safety constraints, human evaluation, or monitoring, especially in user-facing or high-stakes settings. Our results demonstrate improved reward-profile balance under the studied rewards, not guaranteed safety or universal alignment.
We do not collect new user data, recruit human participants, or conduct new human annotation. All experiments use existing public research artifacts, and we report only aggregate evaluation results and case studies without releasing potentially sensitive outputs.

\section*{Acknowledgments}
This research was supported by grants from the National Natural Science Foundation of China (U25B2072, 62337001, 62406303) and Meituan.


\bibliography{custom}

\clearpage

\appendix
\section{Additional Experimental Results and Analysis}
\label{app:additional_analysis}

\subsection{Reward Dynamics}
\label{app:cross_dynamics}

We provide reward and projection-weight dynamics for grounded generation and open-ended alignment.
These settings exhibit reward-profile imbalance similar to that observed in mathematical reasoning: some reward dimensions improve more rapidly or saturate earlier than others, so a fixed scalar projection can over-emphasize dimensions that are easier to optimize.
AMRP mitigates this issue by adapting projection weights online, increasing pressure on reward dimensions that remain lagging, unstable, or stagnant while reducing pressure on dimensions that have already saturated.

\begin{figure}[t]
\centering
\includegraphics[width=0.98\linewidth]{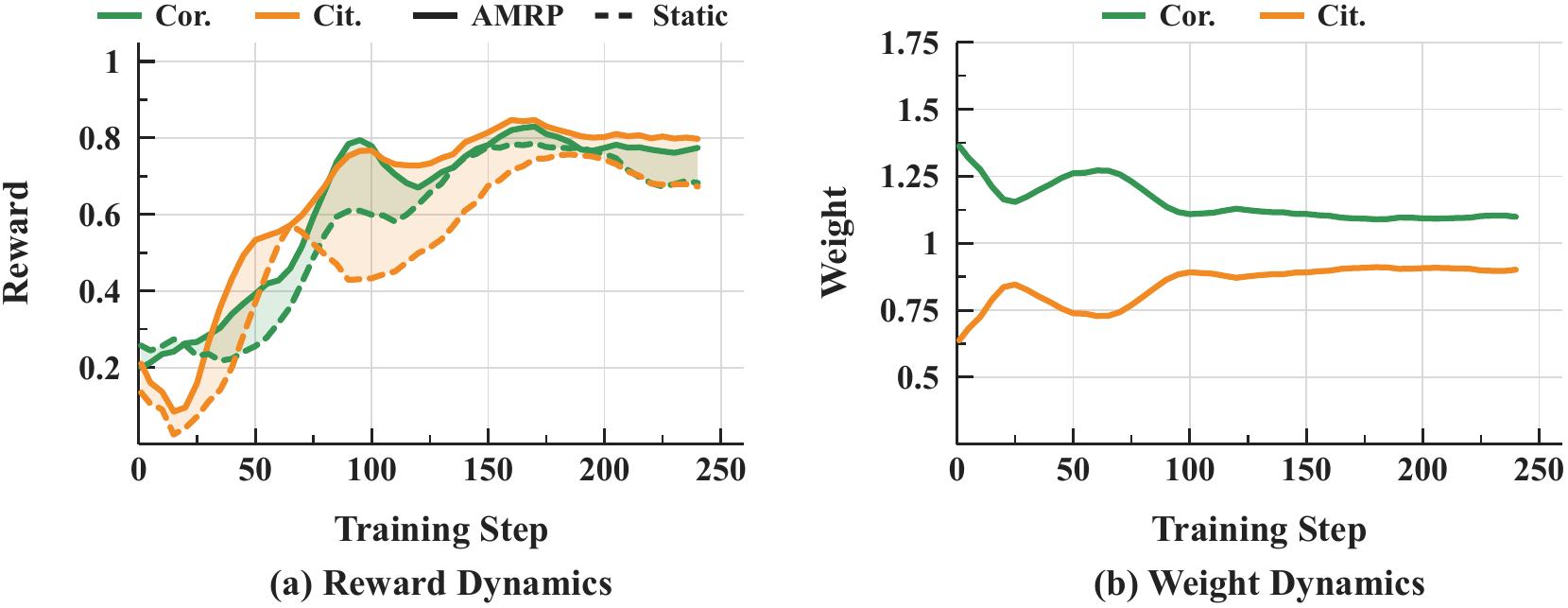}
\caption{
Reward and projection-weight dynamics on ELI5 with Llama-3.1-8B-Instruct.
Under static aggregation, citation support can dominate correctness; AMRP adaptively rebalances the two reward dimensions to mitigate this citation-dominant shortcut.
}
\label{fig:training_dynamics_citation}
\end{figure}

\begin{figure}[t]
\centering
\includegraphics[width=\linewidth]{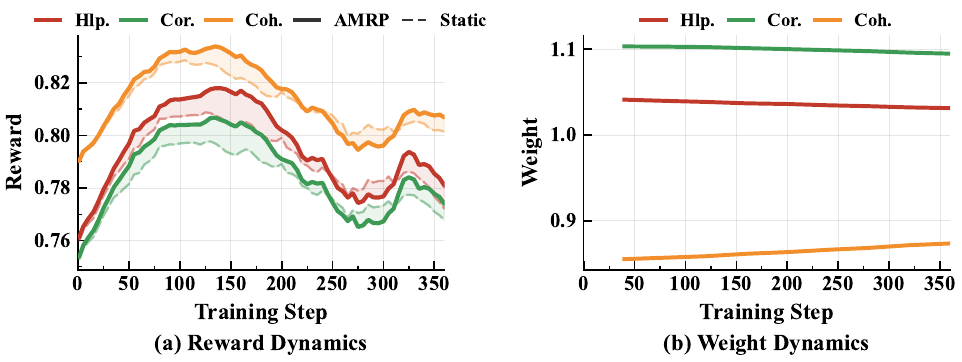}
\caption{Reward and projection-weight dynamics in open-ended alignment with Qwen3-4B-Instruct. AMRP adjusts the projection according to persistent gaps among helpfulness, correctness, and coherence, improving reward-profile balance.}
\label{fig:training_dynamics_helpsteer}
\end{figure}

Figure~\ref{fig:training_dynamics_citation} shows the dynamics for grounded generation.
Citation support improves rapidly under static aggregation, which can make citation rewards dominate the scalar training signal even when answer correctness remains under-optimized.
AMRP reallocates projection weight toward dimensions with larger gaps, leading to a more balanced correctness--citation profile.
Figure~\ref{fig:training_dynamics_helpsteer} shows the dynamics for open-ended alignment.
Although helpfulness, correctness, and coherence are continuous reward-model scores and are more correlated than rule-based rewards, they still exhibit persistent gaps.
AMRP adjusts the projection according to these gaps and improves reward-profile balance under learned reward-model feedback.

\subsection{Qualitative Case Study}
\label{app:case_study}

\begin{figure}[t]
\centering
\includegraphics[width=\linewidth]{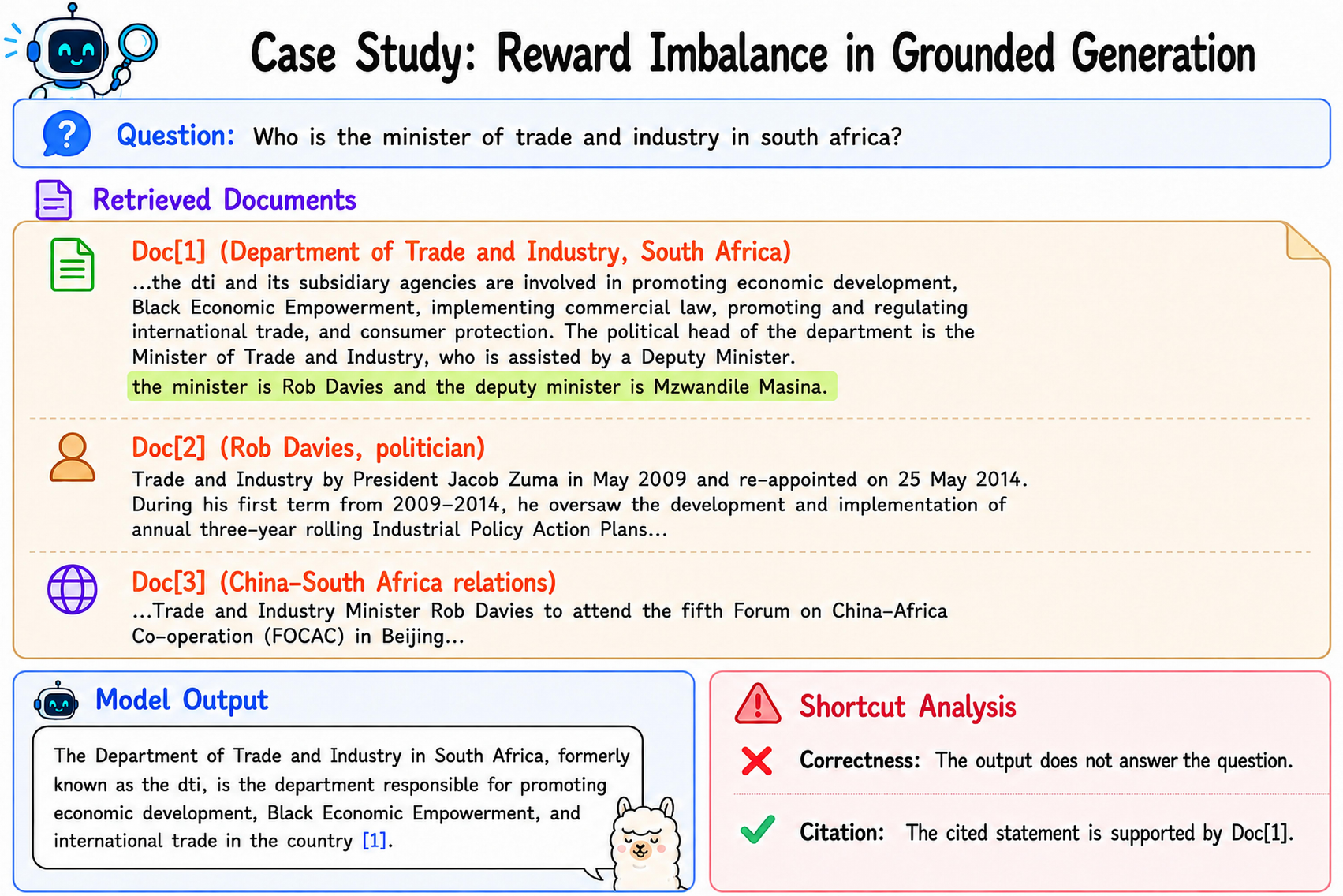}
\caption{
Qualitative case study of a citation-dominant shortcut in grounded generation.
The retrieved evidence identifies the correct answer as ``Rob Davies,'' but the model produces a fluent, cited description of the Department of Trade and Industry without answering the question.
This illustrates aggregation-induced reward hacking, where citation support is satisfied while answer correctness fails.
}
\label{fig:case_study_grounded_generation}
\end{figure}

We further provide a representative qualitative case study of reward imbalance in grounded generation.
As shown in Figure~\ref{fig:case_study_grounded_generation}, the retrieved evidence contains the correct answer, ``Rob Davies,'' and the model output cites the relevant document.
However, the response does not state the correct answer; instead, it produces a fluent description of the Department of Trade and Industry.
This reflects a citation-dominant shortcut: retrieving and citing relevant evidence is easier than performing the precise extraction needed for answer correctness.
Under static aggregation, such responses can receive strong citation rewards while correctness remains under-optimized, illustrating aggregation-induced reward hacking.

\begin{table}[t]
\centering
\small
\setlength{\tabcolsep}{3pt}
\renewcommand{\arraystretch}{1.08}
\resizebox{\columnwidth}{!}{%
\begin{tabular}{llcccccc}
\toprule
\textbf{Param.} & \textbf{Value}
& \textbf{Math} & \textbf{AMC} & \textbf{AIME}
& \textbf{Acc.} & \textbf{Fmt.} & \textbf{Len.} \\
\midrule

\multirow{5}{*}{$\lambda$}
& 0.1 & \textbf{86.80} & 58.40 & 20.94 & 55.38 & 65.90 & 64.73 \\
& 0.5 & 85.80 & 54.22 & 23.33 & 54.45 & 64.91 & 64.71 \\
& 0.7 & 84.80 & 61.11 & 26.56 & 57.49 & 67.37 & 66.72 \\
& 1.0 & 86.60 & \textbf{63.59} & \textbf{31.98}
& \textbf{60.72} & \textbf{70.19} & \textbf{69.34} \\
& 2.0 & \textbf{86.80} & 61.78 & 22.40 & 56.99 & 63.48 & 62.28 \\

\specialrule{\lightrulewidth}{1pt}{1pt}

\multirow{3}{*}{$k_s$}
& 1 & 85.40 & \textbf{65.06} & 20.00 & 56.82 & 68.19 & 67.03 \\
& 5 & \textbf{86.60} & 63.59 & \textbf{31.98}
& \textbf{60.72} & 70.19 & 69.34 \\
& 10 & 85.60 & 61.30 & 26.46 & 57.78 & \textbf{72.90} & \textbf{72.07} \\

\specialrule{\lightrulewidth}{1pt}{1pt}

\multirow{3}{*}{$k_p$}
& 1 & 85.00 & 61.03 & 26.46 & 57.50 & 65.10 & 64.33 \\
& 5 & \textbf{86.60} & 63.59 & \textbf{31.98}
& \textbf{60.72} & \textbf{70.19} & 69.34 \\
& 10 & 86.20 & \textbf{65.36} & 25.31
& 58.96 & 70.15 & \textbf{69.51} \\

\specialrule{\lightrulewidth}{1pt}{1pt}

\multirow{4}{*}{$N$}
& 1 & \textbf{86.60} & 63.59 & 31.98
& \textbf{60.72} & 70.19 & 69.34 \\
& 2 & 86.00 & 57.94 & \textbf{33.33}
& 59.09 & 67.62 & 65.46 \\
& 4 & 86.00 & \textbf{64.31} & 24.69
& 58.33 & \textbf{71.80} & \textbf{71.05} \\
& 8 & 86.00 & 61.63 & 28.44
& 58.69 & 70.37 & 69.30 \\

\bottomrule
\end{tabular}
}
\caption{
Hyperparameter sensitivity of AMRP on Qwen3-4B-Instruct mathematical reasoning.
Each sweep varies one parameter from the default
$(\lambda,k_s,k_p,N)=(1.0,5,5,1)$.
Math, AMC, and AIME report accuracy;
Acc., Fmt., and Len.\ are macro-averages across benchmarks.
}
\label{tab:sensitivity_full}
\end{table}

\begin{table*}[t]
\centering
\small
\setlength{\tabcolsep}{4pt}
\renewcommand{\arraystretch}{1.1}
\resizebox{\linewidth}{!}{
\begin{tabular}{lccc}
\toprule
\textbf{Setting} & \textbf{Training data} & \textbf{Train size} & \textbf{Evaluation data} \\
\midrule
Math reasoning
& NuminaMath-1.5 subset
& 10{,}000
& MATH-500 (500), AMC (2656), AIME 2024 (960) \\
ASQA
& ALCE-ASQA
& 758
& ASQA held-out split (190) \\
ELI5
& ALCE-ELI5
& 800
& ELI5 held-out split (200) \\
Open-ended alignment
& HelpSteer2
& 10{,}161
& AlpacaEval (805), ArenaHard (500), MT-Bench (80) \\
\bottomrule
\end{tabular}
}
\caption{
Dataset statistics for training and evaluation.
Train size denotes the number of prompts used for GRPO fine-tuning.
For AMC and AIME 2024, each unique problem is replicated 32 times in the evaluation data, and metrics are averaged over all generated responses.
}
\label{tab:dataset_stats}
\end{table*}

\begin{table}[t]
\centering
\setlength{\tabcolsep}{3.0pt}
\renewcommand{\arraystretch}{1.08}
\resizebox{\linewidth}{!}{%
\begin{tabular}{lcccc}
\toprule
\textbf{Param.} & \textbf{Math} & \textbf{ASQA} & \textbf{ELI5} & \textbf{HelpSteer2} \\
\midrule
LR                         & $10^{-5}$ & $10^{-5}$ & $10^{-5}$ & $10^{-5}$ \\
Group size $G$             & 8 & 8 & 8 & 8 \\
Steps/gen.                 & 4 & 4 & 4 & 4 \\
Per-dev. batch             & $\{4,8\}$ & 16 & 16 & 4 \\
Max comp. len.             & 3000 & 400 & 500 & 512 \\
Epochs                     & 1 & 3 & 3 & 3 \\
Sampling temp.             & 1.0 & 1.0 & 1.0 & 1.0 \\
\midrule
$\lambda$                  & 1.0 & 0.7 & 0.5 & 0.5 \\
$k_s$                      & 5.0 & 7.0 & 10.0 & 5.0 \\
$k_p$                      & 5.0 & 5.0 & 5.0 & 10.0 \\
$\epsilon$                 & $10^{-3}$ & $10^{-3}$ & $10^{-3}$ & $10^{-3}$ \\
EMA $\rho$                 & 0.9 & 0.9 & 0.9 & 0.8 \\
Update interval $N$        & 1 & 1 & 1 & 10 \\
\bottomrule
\end{tabular}%
}
\caption{
Implementation details for all settings.
The upper block lists GRPO hyperparameters and the lower block AMRP configurations.
Values in braces indicate model-dependent settings, e.g., \(\{4,8\}\) for DeepSeek-Math-7B and Qwen3-4B in math reasoning.
}
\label{tab:impl_details}
\end{table}

\subsection{Hyperparameter Sensitivity}
\label{app:hyperparameter_sensitivity}

We provide full results for the sensitivity analysis in Section~\ref{sec:hyperparameter_sensitivity}, using Qwen3-4B-Instruct in mathematical reasoning. Starting from $(\lambda,k_s,k_p,N)=(1.0,5,5,1)$, we vary one hyperparameter at a time while keeping the others fixed.
Table~\ref{tab:sensitivity_full} reports results for $\lambda$, $k_s$, $k_p$, and $N$. Across all configurations, average accuracy ranges from 54.45 to 60.72, consistently exceeding static aggregation at 37.53 and DRBO$_{\delta}$ at 51.15, both on average and individually on Math, AMC, and AIME. Even the weakest configuration ($\lambda=0.5$, 54.45) remains 3.30 points above DRBO$_{\delta}$. These results show that AMRP does not rely on narrow hyperparameter tuning.
For the update interval $N\in\{1,2,4,8\}$, average accuracy varies only from 58.33 to 60.72, and all settings remain well above both baselines. $N=1$ achieves the highest overall accuracy, suggesting that frequent updates help AMRP respond promptly to reward dynamics. However, performance does not degrade monotonically as $N$ increases: $N=2$ achieves the highest AIME accuracy, while $N=4$ achieves the highest AMC accuracy. Overall, AMRP remains robust to the exact update interval over the tested range.

\section{Additional Implementation Details}
\label{app:implementation}
This appendix provides implementation details for the three experimental settings studied in the main text: mathematical reasoning, citation-grounded long-form QA, and open-ended alignment. We first summarize the datasets and evaluation protocols, then define the reward dimensions used for policy optimization, and finally report the shared GRPO hyperparameters and AMRP-specific configurations.

\subsection{Datasets and Evaluation Protocols}
\label{app:datasets_protocols}

Table~\ref{tab:dataset_stats} summarizes the training and evaluation data used in our experiments.

\paragraph{Mathematical Reasoning.}
We train on a 10K-prompt subset of NuminaMath-1.5 and evaluate on MATH-500, AMC, and AIME 2024. MATH-500 contains 500 unique problems. The AMC and AIME 2024 evaluation sets contain 83 and 30 unique problems, respectively, with each problem replicated 32 times, resulting in 2,656 and 960 evaluation instances. We generate one response for each instance using greedy decoding with temperature 0 and a maximum completion length of 3,000 tokens. All methods are evaluated under the same decoding and replicated-record protocol.

\paragraph{Citation-grounded Long-form QA.}
We follow the ALCE-style grounded-generation setup on ASQA and ELI5.
For each dataset, we use an 80/20 train--test split, resulting in 758 training and 190 held-out ASQA examples, and 800 training and 200 held-out ELI5 examples.
Each prompt contains three retrieved passages and one in-context demonstration.
Responses are sampled with temperature $1.0$, with maximum generation lengths of $400$ tokens for ASQA and $500$ tokens for ELI5.
We exclude ALCE's fluency metric, MAUVE, from the reward profile because it is distribution-level and does not naturally provide a per-response reward for online RL.

\paragraph{Open-ended Alignment.}
We train on HelpSteer2 and evaluate on held-out prompts from AlpacaEval, ArenaHard, and MT-Bench.
Responses are generated greedily with temperature $0$ and a maximum generation length of $512$ tokens.
We report ArmoRM attribute scores scaled to $[0,100]$ rather than official benchmark win rates.

\subsection{Reward Definitions}
\label{app:reward_definitions}

All reward dimensions are oriented such that larger values indicate better responses and are defined or scaled to lie in the $[0,1]$ range. 

\paragraph{Mathematical reasoning rewards.}
For mathematical reasoning, each response is evaluated with
\(\mathbf r_{\mathrm{math}}=[r_{\mathrm{acc}},r_{\mathrm{fmt}},r_{\mathrm{len}}]\).
The accuracy reward \(r_{\mathrm{acc}}\) checks whether the extracted final answer matches the reference answer using exact or symbolic equivalence.
The format reward \(r_{\mathrm{fmt}}\) is binary: it equals \(1\) if the entire response consists of a
\texttt{<reasoning>...\allowbreak</reasoning>} block followed by an
\texttt{<answer>...\allowbreak</answer>} block, allowing only whitespace outside and between the two blocks, and equals \(0\) otherwise.
The length reward softly penalizes overly long generations:
\[
r_{\mathrm{len}}
=
\max\left(
1-\frac{\max(L-2800,0)}{200},
0
\right),
\]
where \(L\) is the number of generated tokens.
Responses with at most \(2{,}800\) tokens receive full reward, which decreases linearly to \(0\) at \(3{,}000\) tokens.

\paragraph{Citation-grounded QA Rewards.}
For ASQA and ELI5, the reward profile is
$r_{\mathrm{alce}}=[r_{\mathrm{corr}},r_{\mathrm{cite}}]$.
For ASQA, $r_{\mathrm{corr}}$ is short-answer exact-match coverage (STR-EM); for ELI5, it is Claims-NLI.
The citation reward $r_{\mathrm{cite}}$ is the F1 score of AutoAIS citation recall and precision using an NLI verifier~\citep{honovich2022true}.

\paragraph{Open-ended Alignment Rewards.}
For open-ended alignment,
$r_{\mathrm{align}}=[r_{\mathrm{help}},r_{\mathrm{corr}},r_{\mathrm{coh}}]$,
corresponding to helpfulness, correctness, and coherence scores from ArmoRM.

\subsection{Training Configurations}
\label{app:training_configurations}

Table~\ref{tab:impl_details} summarizes the shared GRPO hyperparameters and AMRP-specific configurations. All methods within the same experimental setting use identical data, prompts, reward functions, rollout configurations, and optimization hyperparameters. They differ only in how the reward profile is projected into a scalar training signal.
Unless otherwise specified, static aggregation uses mean-one weights $w_i=1$, and AMRP initializes $w_i^{(0)}=1$ for all reward dimensions. For AMRP, projection weights are updated online according to the reward statistics described in Section~\ref{sec:method}.
We fix random seeds for data sampling, rollout generation, and training initialization whenever applicable. Unless otherwise specified, the main tables report single-run results with fixed random seeds. 

\subsection{Artifact Use}
\label{app:artifact_use}

We use existing publicly available research artifacts, including datasets, pretrained models, reward models, benchmarks, and automatic evaluators, only for research purposes. We cite the original creators where the artifacts are introduced. We do not redistribute the original datasets, model checkpoints, or benchmark data. Our released code and artifacts will follow the corresponding licenses, model cards, dataset cards, and terms of use.

\end{document}